\documentclass[sigconf]{acmart}

\AtBeginDocument{%
  \providecommand\BibTeX{{%
    \normalfont B\kern-0.5em{\scshape i\kern-0.25em b}\kern-0.8em\TeX}}}

\copyrightyear{2024}
\acmYear{2024}
\setcopyright{rightsretained}
\acmConference[MM '24]{Proceedings of the 32nd ACM International Conference on Multimedia}{October 28-November 1, 2024}{Melbourne, VIC, Australia}
\acmBooktitle{Proceedings of the 32nd ACM International Conference on Multimedia (MM '24), October 28-November 1, 2024, Melbourne, VIC, Australia}
\acmDOI{10.1145/3664647.3689177}
\acmISBN{979-8-4007-0686-8/24/10}

\usepackage{enumitem}
\usepackage{balance} 

\begin{document}

\title{Multimodal Large Language Models and Tunings:\\Vision, Language, Sensors, Audio, and Beyond}


\author{Soyeon Caren Han}
\affiliation{University of Melbourne \\ \country{Australia}}
\email{caren.han@unimelb.edu.au}

\author{Feiqi Cao}
\affiliation{University of Sydney \\ \country{Australia}}
\email{fcao0492@uni.sydney.edu.au}

\author{Josiah Poon}
\affiliation{University of Sydney \\ \country{Australia}}
\email{josiah.poon@sydney.edu.au}

\author{Roberto Navigli}
\affiliation{Sapienza University of Rome \\ \country{Italy}}
\email{navigli@diag.uniroma1.it}


\begin{abstract}
This tutorial explores recent advancements in multimodal pretrained and large models, capable of integrating and processing diverse data forms such as text, images, audio, and video. Participants will gain an understanding of the foundational concepts of multimodality, the evolution of multimodal research, and the key technical challenges addressed by these models. We will cover the latest multimodal datasets and pretrained models, including those beyond vision and language. Additionally, the tutorial will delve into the intricacies of multimodal large models and instruction tuning strategies to optimise performance for specific tasks.
Hands-on laboratories will offer practical experience with state-of-the-art multimodal models, demonstrating real-world applications like visual storytelling and visual question answering. This tutorial aims to equip researchers, practitioners, and newcomers with the knowledge and skills to leverage multimodal AI. ACM Multimedia 2024 is the ideal venue for this tutorial, aligning perfectly with our goal of understanding multimodal pretrained and large language models, and their tuning mechanisms.
\end{abstract}



\keywords{Multimodal Pretrained Models. Multimodal Large Language Models, Multimodal Tuning, Instruction Tunings}



\maketitle

\section{Tutorial Description and Length}
The proposed tutorial will be a half-day event. The detailed breakdown schedule is as follows:

\begin{enumerate}
    \item \textbf{Introduction (20 mins)}
    \begin{itemize}
        \item What is multimodal?
        \item Historical views and multimodal research tasks
        \item Core Technical challenges
    \end{itemize}

    \item \textbf{Multimodal Pretrained Models (30 mins)}
    \begin{itemize}
        \item Vision Language Datasets
        \item Vision Language Pretrained Models
        \item Other modality Pretrained Models
    \end{itemize}

    \item \textbf{Multimodal Large Models (30 mins)}
    \begin{itemize}
        \item Vision Language Large Models
        \item Other multimodality Large Models
    \end{itemize}

    \item \textbf{Multimodal Instruction Tunings (40 mins)}
    \begin{itemize}
        \item Multi-modality Instruction Tuning Models
        \item Domain specific Instruction Tuning Models
        \item Efficient Fine-Tuning Techniques
    \end{itemize}

    \item \textbf{Break (30 mins)}

    \item \textbf{Demo (60 mins)}
    \begin{itemize}
        \item VLPM downstream task demo (Visual Storytelling, VQA)
        \item VLLM instruction tuning demo
    \end{itemize}

    \item \textbf{Summary and Future Direction (15 mins)}
    \begin{itemize}
        \item Conclusion and Limitation
        \item Future Direction and Trend
    \end{itemize}
\end{enumerate}

\section{Detailed Outline of the Tutorial}
\textbf{Introduction: }
At the beginning of the tutorial on multimodal pretrained and large models, we will briefly introduce the concept of multimodality, including the historical background and general multimodal research tasks. Then, we will dive into the core techniques and challenges addressing the multimodal tasks, which will serve as the foundation for better understanding the development of pretrained models and large models in the following sections.

\noindent\textbf{Multimodal Pretrained Models: }
We will introduce the details of multimodal pretraining, mainly focusing on two modalities, vision and language \cite{ijcai2022p773}. Firstly we will introduce the vision language datasets which are used for pretraining and downstream evaluation, which is related to the potential challenges that researchers face in this domain. We will then discuss the details of different types of vision and language pretrained models, including the general structure and how they are pretrained and finetuned.  We will also cover some pretrained models with other modalities beyond only vision and language, including audio, video, and time-series data.

\noindent\textbf{Multimodal Large Models: }
After the release of ChatGPT, the boom of Large Language Models (LLMs) led to the development of multimodal large models such as BLIP-2~\cite{li2023blip2},  LLaVA~\cite{liu2023llava}, and GPT4V~\cite{gpt4v} which can deal with different modalities as input with the help of LLMs. We will mainly go through the mainstream multimodal large models which can interpret images and perform various downstream vision-language tasks. we will discuss how they are pretrained and what they are normally used for. We will then introduce several models which can understand audio, video and 3D data as well.

\noindent\textbf{Multimodal Instruction Tunings: }
To utilise the power of multimodal large models for diverse downstream tasks, we need to understand the tuning strategy to adapt those models to specific tasks, therefore in this section, we will go through different instruction tuning models such as InstructPix2Pix~\cite{brooks2023instructpix2pix}, LLaVA~\cite{liu2023llava}, InstructBLIP~\cite{dai2023instructblip} and VideoLLaMA~\cite{zhang-etal-2023-video}. Then we will go through some instruction tuning models developed for specific domains such as medical, writing, code generation, sentiment analysis, arithmetic and information retrieval. Since the computational cost of using multimodal large language models is high, we will also discuss several efficient tuning strategies for those models, including LoRA~\cite{hu2022lora} and QLoRA~\cite{dettmers2023qlora}. This will greatly equip the audience with practical techniques for using multimodal large models.

\noindent\textbf{Hands-on Laboratories: }
This tutorial will include two hands-on practical laboratories for using multimodal pretrained models and large models. We will firstly demonstrate how to perform downstream vision-language tasks such as Visual Story Telling (VST) and Visual Question Answering (VQA) on the SOTA multimodal pretrained models introduced before. Then the second demonstration will focus on the instruction tuning for multimodal large models which can be very useful across different domains and real-world applications nowadays. The hands-on laboratories are closely aligned with the content in the previous sessions so that the audience can be equipped with both a theoretical understanding of the multimodal models, and the techniques to use them in practical situations.

\noindent\textbf{Summary and Future Direction: }
In the end, we will briefly summarise the content of the whole tutorial. Due to the seemingly powerful capability of the multimodal pretrained models and multimodal large models, we will also point out the limitations of those models in order to avoid misuse and over-trust on them. In addition, we will provide some insights for discussing the future trend and direction and encourage more researchers to actively address the potential issues of the multimodal models in the near future.

\section{Related Tutorials and Why ACM MM?}

Five tutorials on the related topics were held previously. The first one is \textit{Vision-Language Pretraining: Current Trends and the Future}, held at ACL 2022, introducing the general concept of multimodal pretraining at the early stage. The second tutorial is \textit{Recent Advances in Vision-and-Language Pre-training} held at CVPR 2022, focusing on discussing the multimodal pretrained models from the perspective of the computer vision domain for various specific visual tasks. The third tutorial is \textit{Knowledge-Driven Vision-Language Pretraining} held at AAAI 2023 and CVPR 2023. This tutorial mainly focuses on the injection of knowledge into multimodal pretrained models. In addition, there was one tutorial on transformer structure, namely \textit{Everything You Need to Know about Transformers: Architectures, Optimization, Applications, and Interpretation}, held at AAAI 2023. It briefly discussed multimodal pretraining as one of the subtopics but only focused on the reasoning ability of the models. The last tutorial is \textit{Recent Advances in Vision Foundation Models}, held at CVPR 2024. It covers some discussion for multimodal LLMa but only provides the perspectives from the visual understanding domain. 

All the related tutorials provide limited discussion about only vision-language multimodal pretraining, or only focus on a specific domain of the multimodal models, despite their capability to be utilised across more diverse domains. Our tutorial aims to provide a comprehensive understanding of the development of multimodal (beyond vision-language) pretraining to recent large models with a broad range of domains, from both theoretical and practical points of view. As one of the top-tier venues for Artificial Intelligence and Deep Learning, ACM MM provides a great platform for the discussion of multimodality for both academia and industry, where our tutorial can fill the gap and encourage more valuable work related to multimodality, integrating diverse data types, such as text, images, audio, and video.

\section{Potential Target Audience}
We expect the audience to have a foundational understanding of deep learning and familiarity with commonly-used multimodal tasks. We also welcome newcomers and researchers, offering enriching insights into multimodal deep learning technologies and applying Multimodal LLMs and tuning methods. Through comprehensive lectures and practical laboratories, the tutorial aims to equip attendees with the latest innovations in Multimodal LLMs, ensuring a well-rounded grasp across all skill levels in the field. The \textbf{estimated audience size} would be 20 to 40 people. 

\section{Presenters}
The detailed information of the presenters can be found in Github \url{https://github.com/adlnlp/MultimodalLLM}

\bibliographystyle{ACM-Reference-Format}
\balance
\bibliography{sample-base}










\end{document}